%% file: iclr2027_conference.tex
\documentclass{article} %
\usepackage{iclr2027_conference,times}

\input{math_commands.tex}

\usepackage{hyperref}
\usepackage{url}
\usepackage{graphicx}
\usepackage{booktabs}
\usepackage{caption}
\usepackage{xcolor}
\usepackage{colortbl}
\usepackage{tabularx}
\usepackage[T1]{fontenc}
\usepackage{longtable, array}
\usepackage{courier} %

\definecolor{reported}{gray}{0.45}
\definecolor{evalblue}{RGB}{235,245,253} %
\definecolor{savinggreen}{HTML}{28734F}

\definecolor{casegreen}{HTML}{28734F}
\definecolor{casered}{HTML}{AE403B}
\providecommand{\casereuse}[1]{\textcolor{casegreen}{#1}}
\providecommand{\caserecover}[1]{\textcolor{casered}{#1}}
\providecommand{\PromptFont}{\fontfamily{pcr}\fontsize{8}{9.6}\selectfont}

\title{Pack Light, CarryOn: Building Object-Oriented Context for Long-Horizon Agents}

\author{Yuchen Lu$^{1}$, Julian McAuley$^{1}$, Lei Zhang$^{1}$\thanks{ Corresponding Author.} \\
$^{1}$ University of California San Diego \\
\texttt{yuchlu17@gmail.com \quad jmcauley@ucsd.edu \quad lez023@ucsd.edu}
}

\iclrfinalcopy %
\begin{document}

\maketitle

\begin{abstract}

Long-horizon agents accumulate progress in persistent task state, yet repeatedly process interaction histories describing how that progress was achieved. This raises a natural question: can long-horizon context instead be organized around reusable progress rather than the history that produced it? In this paper, we introduce CarryOn, an \textbf{object-oriented context runtime} that organizes continuation around reusable task objects. Rather than treating compressed histories or reconstructed accounts of progress as the primary basis for continuation, CarryOn builds a context view from continuation carriers, complementary execution context, and a live tail of recent interactions. A core challenge is keeping this reusable basis valid after its surrounding history is retired: how can CarryOn update the basis without reconstructing the full representation at every fold? CarryOn maintains carrier records across folds, revising affected records against bound evidence and applicable environment checks while carrying forward unaffected records and jointly updating execution context. Experiments on OSWorld2.0 and BrowseComp-Plus show improved gains of 5.8\% and 7.0\% while reducing total token usage by 51.9\% and 22.9\%, respectively.
Project page:
\url{https://ethannju.github.io/carryon}.

\end{abstract}

\input{sec/1_intro}
\input{sec/2_related}
\input{sec/3_method}

\input{sec/4_experiment}
\input{sec/5_conclusion}

\bibliography{iclr2027_conference}
\bibliographystyle{iclr2027_conference}

\newpage
\appendix
\section{Method and Implementation Details}

\subsection{Context Representation and Update Procedure}
\label{app:Context Representation and Update Procedure}
CarryOn maintains an append-only interaction log, a candidate registry,
versioned carrier records, and execution-context revisions.
The committed carrier records and execution context define the folded
portion of the context view; subsequent interactions form the live tail.

\paragraph{Candidate Records.}
Each candidate has a runtime-assigned identifier, a provisional carrier
type, a short description, and references to the interactions supporting
it. Its identity fields specify the associated object or subject, and it
may refer to an existing carrier whose progress is being updated.
The registry records when the candidate was first observed and last
revised. Candidates can persist across Tracker invocations, allowing
progress to accumulate before compilation. Registering or revising a
candidate does not itself trigger a fold.

\paragraph{Carrier Records and Versions.}
A committed carrier record contains its identity, carrier type, reusable
content or object reference, conditions for continuation, applicability
scope, provenance, and version information. External-object identities
are derived from their normalized locations. Information-product
identities combine a representation type with a structured subject,
while situated-state identities combine a state domain with the subject
of an environment anchor. These identity fields distinguish the object
being tracked from its changing contents.

Updating an existing carrier creates a successor record linked to the
previous version. The successor becomes current, while earlier versions
remain stored for provenance. Carriers not changed by a fold retain their
current records. For information products, reusable content is stored as
claims with identifiers, propositions, optional values, and supporting
source references. Revisions can add, modify, or retract claims while
carrying forward unaffected claims.

\paragraph{Execution Context.}
Execution context is stored as a revision associated with a committed
fold. It contains the current focus, continuity information needed to
resume work, open obligations, active constraints, and immediate
continuation instructions. Entries retain source attribution.
The Compiler revises this information against the proposed carrier
updates, preserving unresolved requirements and removing items that
have been resolved or are now represented by carriers.

\paragraph{Update Procedure.}
At each invocation, the runtime performs the following operations:
\begin{enumerate}
    \item The Tracker reads recent interactions and the candidate registry.
    The runtime applies admissible candidate updates and resolves any
    compilation requests to their candidate records.

    \item For each admitted request, the runtime binds the candidate
    snapshot, authorized prior carrier records, and supporting sources.
    Applicable environment checks contribute additional evidence.
    The resulting package fixes the records and evidence used for
    compilation. If no request is admitted, the invocation ends without
    a fold.

    \item The Compiler proposes carrier additions or revisions together
    with a new execution context. Its input also includes the current
    execution context and a chronological text projection of the live
    tail, so continuation information outside the selected candidates
    can be retained.

    \item The runtime checks identity and reference consistency, source
    authorization, output constraints, and whether the compilation input
    records remain current. A valid proposal must contain an actual carrier
    change. Acceptance atomically commits the carrier updates and
    execution-context revision and advances the fold boundary through
    the entire live tail. Rejection leaves these committed components
    and the live tail unchanged.
\end{enumerate}

Previously accepted candidate updates remain available after a rejected
compilation proposal. The original interactions remain in the log after
a successful fold. Grounder rules and installation of the committed view
are described in Appendices~\ref{app:grounder}
and~\ref{app:implementation}, respectively.

\subsection{Grounder Checks and Verdict Rules}
\label{app:grounder}
CarryOn aligns compilation requests with their supporting evidence.
Information products undergo source-reference and excerpt checks.
For object-bound requests, the Grounder performs scoped, read-only
environment checks. Table~\ref{tab:grounder-checks} summarizes the
implemented checks in the OSWorld 2.0 integration.

\begin{table}[t]
\centering
\caption{Read-only environment checks in the OSWorld 2.0 integration.}
\label{tab:grounder-checks}
\small
\renewcommand{\arraystretch}{1.15}
\setlength{\tabcolsep}{5pt}
\begin{tabularx}{\linewidth}{
@{}>{\raggedright\arraybackslash}p{0.23\linewidth}
>{\raggedright\arraybackslash}X@{}
}
\toprule
Check & Observation \\
\midrule

File existence
& Whether the file is present at its bound path. \\

Change from initial state
& Whether the file-content hash differs from its registered initial snapshot. \\

File size
& Whether the file is empty or nonempty. \\

Document structure
& Whether a \texttt{docx}, \texttt{pptx}, or \texttt{xlsx} file has a valid document container. \\

Document differences
& Changes in document text units, added or removed parts, and placeholder-like content. \\

Instruction text matching
& Presence or absence of a text span copied from the public task instruction. \\

\bottomrule
\end{tabularx}
\end{table}
\paragraph{Request binding and evidence records.}
Each check is bound to a candidate identity and revision, an object reference, a predicate, and an expected value before execution.
The runtime records the observation and its provenance, and validates
the binding against the current candidate before using the result.

\paragraph{Verdicts and compilation.}
A condition is \texttt{supported} when the observation establishes
the expected value, \texttt{contradicted} when it establishes an
incompatible value, and \texttt{unknown} when the evidence cannot
decide. A request is supported only when all declared conditions
are supported. Any contradiction makes it contradicted.
Otherwise, an unresolved condition makes it unknown.

Supported object-bound requests may proceed to compilation.
Unsupported checks or unknown results leave the affected candidate
uncompiled, while other admissible requests may proceed.
If the interaction evidence supports an independently reusable
information product, the Tracker may register a separate candidate
through the source-validation route.

These checks serve as supporting consistency safeguards within the runtime, helping the agent identify explicit contradictions and outstanding task requirements.

\subsection{Agent Integration and CarryOn Settings}
\label{app:implementation}

\paragraph{Agent integration.}
\label{app:carryonsetting}
CarryOn runs alongside the agent through a benchmark-specific adapter.
The adapter records interactions, supplies their textual representations
and exposed thinking summaries to the runtime, and installs committed
carrier records and execution context before the next agent call.
Within each benchmark, the baseline and CarryOn use the same agent model,
environment-action interface, and task budget.
CarryOn additionally supplies the context-management interfaces and
feedback described below.

\paragraph{Model configuration.}
We use Claude Opus 4.7 for the agent, Tracker, and Compiler, without
additional training.
Table~\ref{tab:implementation-settings} lists their output limits and
execution budgets.
On OSWorld 2.0, the agent uses adaptive thinking at maximum effort;
Tracker and Compiler requests contain no explicit thinking or effort
override.
On BrowseComp-Plus, the agent uses adaptive thinking with the provider's
default effort, the Tracker uses adaptive thinking at high effort,
and the Compiler has thinking disabled.
Both adapters expose the provider's summarized thinking output to the
runtime, marking it as agent-generated information rather than
environment observations.

\begin{table}[t]
\centering
\caption{Agent and CarryOn settings.
Output limits are measured in tokens per model call.}
\label{tab:implementation-settings}
\small
\renewcommand{\arraystretch}{1.12}
\setlength{\tabcolsep}{6pt}
\begin{tabularx}{\linewidth}{@{}Xcc@{}}
\toprule
Setting & OSWorld 2.0 & BrowseComp-Plus \\
\midrule
Agent execution budget & 500 steps & 60 turns \\
Agent output limit & 16,000 & 32,000 \\
Tracker output limit & 2,048 & 8,192 \\
Compiler output limit & 8,192 & 8,192 \\
Compilation requests per scan & 4 & 4 \\
\bottomrule
\end{tabularx}
\end{table}

\paragraph{Runtime triggering.}
On OSWorld 2.0, the Tracker is normally invoked when interaction text
accumulated since the previous scan or fold exceeds an estimated
6,000 tokens.
This estimate is the character count of the runtime's text projection
divided by four and excludes the agent's thinking summaries.
These thresholds measure text volume, not predicted monetary savings.

On BrowseComp-Plus, scan scheduling uses growth in the agent's recorded
prompt size, including cached input.
The initial trigger requires 120,000 tokens of growth plus a
1,024-token margin and newly ingested search results.
Consecutive scans that produce no fold double the token thresholds,
up to six doublings; a successful fold resets this backoff.

\paragraph{OSWorld integration.}
The adapter preserves the initial task message and installs a serialized
checkpoint containing carrier records and execution context.
It retains the contiguous suffix of complete message groups needed to
preserve screenshots kept by the agent's native image filter, together
with the pending assistant message needed for protocol continuity.
The native filter is configured with a recent-image threshold of 10,
and screenshot eviction remains under its control.
Retained tool calls, results, and signed thinking blocks preserve their
protocol relationships.
A fixed system instruction identifies the checkpoint as runtime-supplied
context while treating its contents as task data.

\paragraph{BrowseComp-Plus integration.}
The agent uses BM25 search with up to five results per call and
512-token snippets, without full-document retrieval.
The runtime receives bounded textual projections of search results,
including document identifiers, ranks, scores, and the first
160 characters of each snippet, alongside agent text and exposed
thinking summaries.
After a fold, pending search results remain available to the next
agent call.

\paragraph{Installation and recovery.}
Before installing a new context, the adapter snapshots the agent's
message state.
A failed installation restores this snapshot and the previous
active-checkpoint bookkeeping.
If restoration is verified, execution continues with further runtime
intervention disabled.
If the previous state cannot be verified, the run stops with an error.
This separates commitment of a runtime representation from its
successful installation in the agent's context.

\section{Experimental Details}

\subsection{Benchmark and Task Selection}
\label{app:benchmark-details}
\paragraph{OSWorld 2.0.}
OSWorld 2.0 contains long-horizon tasks spanning seven professional
domains and 21 subcategories \citep{yuan2026osworld20benchmarkingcomputer}.
Its workflows combine desktop applications and web services, requiring
cross-source reasoning, state tracking, and coordinated operations on
multiple task outputs. The benchmark reports a median estimated human
completion time of approximately 1.6 hours.
The duration and diversity of these workflows make it suitable for
evaluating context management over extended task execution.
We use the \texttt{v2026.08.08} release.

Our main OSWorld 2.0 evaluation covers the full evaluation set,
with both the baseline agent and CarryOn independently run by us.
For the additional context-management comparisons and ablation
experiments, we sample a shared subset of 40 tasks,
with the following task IDs:
\begin{quote}
\small
001, 002, 003, 004, 005, 006, 008, 009, 012, 013,
014, 018, 030, 031, 038, 039, 040, 046, 047, 048,
049, 057, 059, 061, 063, 064, 066, 070, 071, 073,
076, 079, 082, 083, 087, 089, 093, 094, 095, 102.
\end{quote}

\paragraph{BrowseComp-Plus.}
BrowseComp-Plus contains 830 challenging questions over a fixed corpus
of 100,195 documents \citep{chen2025browsecompplusfairtransparentevaluation}.
Its questions require iterative search and reasoning across multiple
sources, producing growing histories of retrieved evidence and
intermediate analysis. This setting allows us to evaluate context
management under sustained information accumulation.
We use BM25 retrieval and evaluate both the baseline and CarryOn
on the same 100-question subset, with the following question IDs:
\begin{quote}
\small
11, 12, 15, 18, 36, 39, 209, 223, 228, 231,
233, 237, 240, 241, 246, 249, 251, 255, 256, 264,
266, 268, 480, 484, 485, 487, 495, 496, 500, 501,
502, 505, 509, 513, 515, 517, 519, 527, 528, 529,
530, 537, 544, 546, 549, 553, 555, 562, 770, 771,
773, 774, 776, 784, 787, 793, 796, 798, 801, 802,
805, 809, 819, 820, 821, 822, 827, 828, 833, 844,
850, 852, 856, 863, 869, 872, 873, 986, 991, 998,
1005, 1015, 1016, 1020, 1023, 1025, 1029, 1030, 1033, 1036,
1037, 1038, 1039, 1040, 1042, 1046, 1049, 1060, 1063, 1066.
\end{quote}

\subsection{Previously Reported Results}
\label{app:reported-results}

\paragraph{OSWorld 2.0.}
We include the published 500-step results from the original
OSWorld 2.0 evaluation \citep{yuan2026osworld20benchmarkingcomputer}, retaining its
reported scores and estimated model cost per task.
All models receive screenshot observations through the benchmark's
official agent implementations.
Claude models use their native computer-use interface, whereas all other models generate PyAutoGUI actions. GPT-5.5 executes actions in batches, while the reported Claude, MiniMax, Kimi, and Qwen results use the same single-action setting as our evaluation.

\paragraph{BrowseComp-Plus.}
We compute external reference results on the same 100 questions
used in our experiments.
For GPT-5 and o3, we use the benchmark authors' publicly released
BM25 runs \citep{chen2025browsecompplusfairtransparentevaluation} and grade their available
final responses with Qwen3-32B using the benchmark's official
grading prompt.
For GPT-OSS-120B\citep{gptoss}, we use the BM25\citep{bm25} configuration as the sparse retriever.

\subsection{Evaluation and Resource Accounting}
\label{app:usage-accounting}
\paragraph{Accounting scope.}
We report resource usage as per-task averages.
For each selected task attempt, we accumulate recorded usage
from both the agent and auxiliary context-management calls.
For CarryOn, the latter includes the Tracker and Compiler.
For summarization-based methods, it includes summarizer calls.
We report agent usage, auxiliary usage, and their combined totals,
so the total captures the resources consumed by both task execution
and context management.

\paragraph{Token accounting.}
Token counts come from API usage records.
For task $i$, let $\mathcal{C}_i$ contain its recorded agent
and context-management calls. Total token usage is
\[
T_i =
\sum_{j \in \mathcal{C}_i}
\left(I_j + R_j + W_j + O_j\right),
\]
where $I_j$ denotes uncached input tokens,
$R_j$ cache-read tokens, $W_j$ cache-write tokens,
and $O_j$ output tokens.
Cache writes include both five-minute and one-hour entries.
When usage records provide an overall cache-write count and
its lifetime breakdown, the breakdown partitions that count
and is not added again.
Reasoning tokens already included in output usage are also
counted once.

Token usage is cumulative across requests: input processed again
in a later call is counted again, including input served from
the cache. This measures the total tokens processed during
execution. We report token usage in millions,
with $1\,\mathrm{M}=10^6$ tokens.

\paragraph{Cost accounting.}
For our Claude Opus 4.7 runs, we calculate nominal API-equivalent
costs using the fixed rates in Table~\ref{tab:accounting-rates}.
Each call's cost is the sum of its token counts multiplied
by the corresponding category rates, and task cost is the
sum across calls. The same rates apply to the agent and
CarryOn's auxiliary models.

\begin{table}[t]
\centering
\caption{Nominal token prices used for resource accounting.}
\label{tab:accounting-rates}
\small
\renewcommand{\arraystretch}{1.12}
\begin{tabular}{lr}
\toprule
Token category & Price (\$/M tokens) \\
\midrule
Uncached input              & 5.00  \\
Cache read                  & 0.50  \\
Cache write (5 minutes)     & 6.25  \\
Cache write (1 hour)        & 10.00 \\
Output                      & 25.00 \\
\bottomrule
\end{tabular}
\end{table}

Cached input remains part of total token usage but receives
its corresponding cache price. Consequently, reductions in
token usage and cost can differ depending on the proportions
of uncached input, cache reads, cache writes, and output.
Reported totals include context-management overhead and
exclude evaluator, user-simulator, and infrastructure costs.
They cover the recorded usage of the selected task attempts.

\paragraph{Aggregation.}
For each method, we average task totals over its reported
task set. Relative reductions in token usage and model cost
are computed from these means:
\[
\operatorname{Reduction}(x)
=
100\left(
1-\frac{\bar{x}_{\mathrm{CarryOn}}}
        {\bar{x}_{\mathrm{baseline}}}
\right)\%,
\]
where $\bar{x}$ is mean token usage or model cost per task.
All calculations use unrounded values, with rounding applied
only to the displayed results.

\section{Supplementary Results and Analysis}
\subsection{Statistical Procedures for Context and Carrier Analysis}
\label{app:context-carrier-statistics}

\paragraph{Context and cost dynamics.}
Figure~\ref{fig:costana} uses the per-call usage traces of all
baseline and CarryOn trajectories included in the resource comparison.
An agent turn denotes a metered agent-model response, including
retries and terminal-review responses with recorded usage.
For method $m$, let $L_{m,i}$ denote the number of such responses
in trajectory $i$, and define the active set at turn $t$ as
$\mathcal{A}_m(t)=\{i:L_{m,i}\geq t\}$. We compute
\[
\bar P_m(t)
=
\frac{1}{|\mathcal A_m(t)|}
\sum_{i\in\mathcal A_m(t)} P_{m,i,t},
\qquad
\bar C_m(t)
=
\frac{1}{|\mathcal A_m(t)|}
\sum_{i\in\mathcal A_m(t)} C_{m,i}(t).
\]
Here, $P_{m,i,t}$ is the input token count reported by the API,
including uncached input, cache reads, and cache creation.
Screenshot tokens remain included.
$C_{m,i}(t)$ accumulates agent costs through turn $t$ and runtime
costs assigned to the same boundary, using the accounting rules
in Appendix~\ref{app:usage-accounting}.
Gray bars show the fraction of baseline trajectories that remain
active at each turn, sampled every ten turns.

\paragraph{Curve smoothing and annotations.}
We first compute the per-turn means and then apply a Gaussian
filter with a standard deviation of 12 turns, a kernel truncated
at four standard deviations, and symmetric reflection at the boundaries.
Input-context curves use the complete recorded series before
the displayed range is restricted to 500 turns.
Cost curves are smoothed over turns 0--500, with the unsmoothed
series initialized at zero cumulative cost. Gray bars are also smoothed for visualization.
Figure annotations are calculated from the raw data.
Shading shows the difference between the two cost curves.

\paragraph{Post-fold text length.}
For Table~\ref{tab:osworld40}, we measure the textual context
immediately after each installed fold, before the next observation
is appended.
All methods use the same text projection, including task instructions,
retained records or summaries, interaction text, returned thinking
summaries, and role/action/observation labels.
Image data, signatures, fixed system prompts, and tool schemas
are excluded.
Length is measured in thousands of Unicode characters.
We first average lengths across folds within each task and then
average equally across tasks with at least one fold.
The baseline has no post-fold measurement.
We use this measure together with task performance to compare how much text each method retains and how well the agent performs.

\paragraph{Carrier persistence across folds.}
For Figure~\ref{fig:ideaana}(a), we compare current carrier records
before and after each committed and installed fold.
We examine all 494 committed and installed folds across the
CarryOn trajectories included in this analysis.
Carrier identities are tracked across record versions using
registered identities and successor relations.
A carrier is \emph{retained} when its current record version is
unchanged, \emph{updated} when a successor version is committed,
and \emph{added} when it has no predecessor.
Each carrier contributes one observation per fold in which it is
present afterward. The same carrier can therefore contribute
across several folds, and initial creations are included.
Pooling these observations gives 278 retained, 372 updated,
and 163 added observations, for 813 in total.
The retained-or-updated proportion is
$650/813 \approx 79.95\%$, rounded to $80\%$ in the figure.
These percentages describe record continuity across folds.

\subsection{Continuation Cases and Ablation Behaviors}
\label{app:case-study}
\label{app:case-study}

We inspect recorded trajectories to show how execution continues after a fold
and how the agent behaves when carrier records are hidden. The tables present
selected steps, the information retained at the fold, and subsequent actions.

\paragraph{Continuing from established progress.}
Table~\ref{tab:case-form} shows a publication form that remains populated across
a fold. The agent checks the existing entries and submits the form without
re-entering the authors. The saved screenshot confirms submission, and the
paper subsequently appears in the Research Output list.

\paragraph{Returning to previously acquired information.}
Table~\ref{tab:case-carrier-ablation} compares two runs of a prize-distribution
task. In the full CarryOn run, the agent continues filling the form after
the fold. In the carrier-masking run, it has already established the item-price
mapping and used it to construct a plan. After the fold, it questions whether
the prices have been confirmed, extracts the PDF again, and reopens the
catalogue to reconstruct the mapping. This provides a concrete example of
repeated information acquisition when carrier records are withheld.

\begin{table}[t]
\centering
\small
\renewcommand{\arraystretch}{1.14}
\setlength{\tabcolsep}{6pt}
\caption{\textbf{Continuing from a populated form (Task 009).} Selected steps from a main-experiment CarryOn trajectory. Green highlights continuation from the existing form. Entries are condensed from the recorded trajectory; the submission confirmation is also visible in the saved screenshot.}
\label{tab:case-form}
\begin{tabularx}{\linewidth}{@{}>{\raggedright\arraybackslash}p{0.12\linewidth} >{\raggedright\arraybackslash}X >{\raggedright\arraybackslash}p{0.25\linewidth}@{}}
\toprule
Step & Recorded state or agent statement & Action \\
\midrule
203 & The agent reports that all four external authors have been entered. & Continues through the form. \\
205 & Publication metadata is entered alongside the author list. & Types the NeurIPS conference dates. \\
F4 after 240 & Carrier: entered authors and publication metadata. EC: classify authors by affiliation at publication time. & Installs the updated context view. \\
247 & \casereuse{Reviews the existing author classifications against the guidelines.} & Presses Escape after the review. \\
249 & The publishing status, peer-review status, and attached PDF remain in the form. & \casereuse{Returns to the form without re-entering the authors.} \\
250 & The existing form is ready for submission. & \casereuse{Clicks Save \& Submit; the page confirms successful submission.} \\
253 & \casereuse{The submitted paper appears in the Research Output list.} & Ends the run. \\
\bottomrule
\end{tabularx}
\vspace{3pt}
\begin{minipage}{\linewidth}\footnotesize The earlier author and conference-entry steps precede the retained interaction window (221--240). The agent also reopens the guidelines after the fold.\end{minipage}
\end{table}

\begin{table}[t]
\centering
\small
\renewcommand{\arraystretch}{1.14}
\setlength{\tabcolsep}{6pt}
\caption{\textbf{Continuing a plan versus reacquiring its inputs (Task 013).} Selected steps from the full CarryOn and carrier-masking runs. Green marks continuation of the retained prize plan; red marks renewed uncertainty and repeated acquisition of the price mapping. Quoted spans are exact excerpts from recorded agent reasoning summaries; other descriptions are condensed.}
\label{tab:case-carrier-ablation}
\begin{tabularx}{\linewidth}{@{}>{\raggedright\arraybackslash}p{0.12\linewidth} >{\raggedright\arraybackslash}X >{\raggedright\arraybackslash}p{0.25\linewidth}@{}}
\toprule
Step & Recorded state or agent statement & Action \\
\midrule
\multicolumn{3}{@{}l}{\textbf{CarryOn}} \\
\addlinespace[3pt]
30 & Identifies all 15 catalogue items and their prices, then develops a prize-distribution plan. & Closes the catalogue viewer. \\
F2 after 57 & Carrier: item-price mapping and chosen tier assignments. EC: enter the third-prize total, then complete the participation tier. & Both records remain visible to the agent. \\
58 & \casereuse{Continues entering the planned third-prize total.} & \casereuse{Types 125 in the total field.} \\
60 & \casereuse{Moves on to the planned participation-prize items.} & \casereuse{Selects Chocolate Box.} \\
\midrule
\multicolumn{3}{@{}l}{\textbf{w/o Carrier}} \\
\addlinespace[3pt]
35 & ``Now I have all the pricing details I need''; selects a plan with tier totals 126/125/125/124. & Begins entering the plan. \\
F2 after 54 & The runtime compiles the price mapping and plan, but carrier records are hidden from the agent. EC retains the tier totals and next entry step. & Installs the masked view. \\
55 & \caserecover{``I don't actually have the item prices confirmed''.} & Types 125 in the total field. \\
56--57 & \caserecover{Returns to ``checking the PDF again for the item-to-price mapping''.} & \caserecover{Opens a terminal.} \\
59--60 & \caserecover{Extracts the source catalogue again.} & \caserecover{Runs pdftotext on prize-items.pdf.} \\
65--69 & \caserecover{Renders the catalogue pages and reopens page 2.} & \caserecover{Runs pdftoppm, then xdg-open.} \\
71--72 & \caserecover{Reads item-price correspondences from page 2 again.} & \caserecover{Zooms in on the reopened catalogue.} \\
\bottomrule
\end{tabularx}
\vspace{3pt}

\end{table}

\section{Prompts}
\label{app:prompts}

The Tracker and Compiler use separate system prompts. The runtime selects
an environment, information-product, or composite Compiler route according to the
authorized candidates. For presentation, prompt terminology is standardized
to match the paper: continuation carriers (carriers for short), the host agent,
and the runtime scheduler. An environment-check specification lists conditions
for the Grounder to check; selected evidence forms part of the compilation
input package $U_t$. Internal policy labels, development annotations, and
serialization-version tags are omitted. JSON field names and categorical
values retain their implementation spelling.

\paragraph{Runtime inputs.}
The Tracker receives the task and current committed records, the live tail,
and a current-state block containing candidates and runtime feedback. The
Compiler additionally receives the authorized compilation requests and their
source bindings. These blocks are populated for each invocation.

\paragraph{Summary inputs.}
GS and CGS receive the current visible conversation, including the task,
any earlier summary, and recent messages. The input preserves message roles,
tool actions, returned observations, visible images, and returned reasoning
summaries where available. They return a replacement summary as plain text.

The following tables present selected instructions from the main system prompts with the editorial changes described above; GS is reproduced in full. The marker \texttt{[...]} indicates omitted passages.

\begingroup
\PromptFont
\setlength{\tabcolsep}{6pt}
\renewcommand{\arraystretch}{0.96}
\setlength{\LTpre}{8pt}
\setlength{\LTpost}{12pt}
\setlength{\LTcapwidth}{\linewidth}
\begin{longtable}{@{}p{\linewidth}@{}}
\caption{\normalfont\small\textbf{Tracker prompt.} Selected instructions from the routine Tracker system prompt, adapted for presentation. Ellipses mark omitted passages.}\label{tab:prompt-tracker}\\
\toprule
\endfirsthead
\multicolumn{1}{@{}l@{}}{\normalfont\small\textbf{Tracker prompt (continued)}}\\
\toprule
\endhead
\midrule
\multicolumn{1}{r}{\normalfont\scriptsize Continued on the next page}\\
\endfoot
\bottomrule
\endlastfoot
\addlinespace[4pt]
{\normalfont\footnotesize\bfseries System prompt}\\*
\addlinespace[3pt]
\mbox{You are the Carrier Tracker and Readiness Judge of a Context Runtime for a}\\
\mbox{computer-use agent. You observe the agent's interaction stream step by step and}\\
\mbox{hold TWO equally weighted duties:}\\
\rule{0pt}{4pt}\\[-4pt]
\mbox{DUTY A \textemdash{} LEDGER: maintain a sparse identity ledger of the continuation carriers}\\
\mbox{the agent is producing (the candidates).}\\
\mbox{DUTY B \textemdash{} READINESS: judge whether a coherent carrier version can be constructed}\\
\mbox{from what the stream already shows, and express that judgment as a}\\
\mbox{compile\_request. The runtime scheduler then authorizes exactly one of two}\\
\mbox{evidence routes before a separate Compiler: environment objects require an}\\
\mbox{aligned typed environment-check specification; information products require a}\\
\mbox{sealed, bounded stream-evidence envelope. The runtime validation and commit}\\
\mbox{logic then decide commit.}\\
\rule{0pt}{4pt}\\[-4pt]
\mbox{\textcolor{gray}{[...]}}\\
\rule{0pt}{4pt}\\[-4pt]
\mbox{Continuation carriers (what the ledger tracks) \textemdash{} a persistent work product later}\\
\mbox{steps can read, modify, or consume WITHOUT replaying the actions that produced}\\
\mbox{it. Classify by WHERE THE USABLE VALUE LIVES and whether it can be consumed}\\
\mbox{independently of the current execution environment \textemdash{} NEVER by whether the agent}\\
\mbox{feels the task or a phase is complete. Incompleteness does not imply State; lack}\\
\mbox{of a pre-existing file does not imply State.}\\
\mbox{- PRODUCT: the semantic content can be externalized and consumed away from the}\\
\mbox{\hspace*{2\fontdimen2\font}current execution process. Test: if this VM, session and trajectory all}\\
\mbox{\hspace*{2\fontdimen2\font}vanished and only this representation survived, could a new agent understand}\\
\mbox{\hspace*{2\fontdimen2\font}and use it? Yes -> product. Half-done but readable documents, organized}\\
\mbox{\hspace*{2\fontdimen2\font}findings and derivations, phase analyses, plans, configurations all qualify \textemdash{}}\\
\mbox{\hspace*{2\fontdimen2\font}intermediate or final. Two carriers:}\\
\mbox{\hspace*{2\fontdimen2\font}* external\_object: the content already lives in an independently addressable}\\
\mbox{\hspace*{4\fontdimen2\font}object (a file, a saved document).}\\
\mbox{\hspace*{2\fontdimen2\font}* information\_product: the content is carried by structured CLAIMS with}\\
\mbox{\hspace*{4\fontdimen2\font}evidence pointers, compiled from the interaction stream; the registry}\\
\mbox{\hspace*{4\fontdimen2\font}externalizes it at commit \textemdash{} no file needed. Pages the agent merely read are}\\
\mbox{\hspace*{4\fontdimen2\font}NOT information products \textemdash{} the task-specific record compiled FROM them is.}\\
\mbox{\hspace*{4\fontdimen2\font}An investigation/repair/analysis workflow gets its information-product}\\
\mbox{\hspace*{4\fontdimen2\font}candidate EARLY \textemdash{} when the agent starts gathering facts, not when the record}\\
\mbox{\hspace*{4\fontdimen2\font}feels finished.}\\
\mbox{- STATE (situated\_state): what must survive is the current environment or}\\
\mbox{\hspace*{2\fontdimen2\font}process configuration ITSELF \textemdash{} language can describe it but not replace it.}\\
\mbox{\hspace*{2\fontdimen2\font}Test: must a new agent return to this same environment, session or process to}\\
\mbox{\hspace*{2\fontdimen2\font}actually continue? Yes -> state (a checkout at the payment step, a game in}\\
\mbox{\hspace*{2\fontdimen2\font}progress, a login session, an unsaved workbook). A state is not "an unfinished}\\
\mbox{\hspace*{2\fontdimen2\font}product", and a product never becomes state just because it is intermediate.}\\
\rule{0pt}{4pt}\\[-4pt]
\mbox{\textcolor{gray}{[...]}}\\
\rule{0pt}{4pt}\\[-4pt]
\mbox{SEMANTIC CARRY RULE:}\\
\mbox{- Optimize useful context per retained token. A wake creates no duty to compile,}\\
\mbox{\hspace*{2\fontdimen2\font}and the examples below are not a noun checklist. First ask what a future host}\\
\mbox{\hspace*{2\fontdimen2\font}agent must know to continue, decide, or avoid material rediscovery after this}\\
\mbox{\hspace*{2\fontdimen2\font}tail is deleted.}\\
\mbox{- Name one stable Candidate for the thing future work consumes. Use}\\
\mbox{\hspace*{2\fontdimen2\font}external\_object for an independently addressable object, information\_product}\\
\mbox{\hspace*{2\fontdimen2\font}for a maintained result/record across sources, and situated\_state only under}\\
\mbox{\hspace*{2\fontdimen2\font}its existing same-environment contract. Target the active Candidate C-id, not}\\
\mbox{\hspace*{2\fontdimen2\font}its parent A-id; inherit the latest parent and never duplicate forms.}\\
\mbox{- Define the Candidate's concrete current version before choosing handles. A}\\
\mbox{\hspace*{2\fontdimen2\font}useful version may be a scoped result, requirement, decision, constraint,}\\
\mbox{\hspace*{2\fontdimen2\font}unresolved question, supported failed route, or recoverable in-flight}\\
\mbox{\hspace*{2\fontdimen2\font}boundary. Retain it only when deletion would force meaningful rereading,}\\
\mbox{\hspace*{2\fontdimen2\font}rechecking, re-decision, retry, or reconstruction. Mere chronology, UI or}\\
\mbox{\hspace*{2\fontdimen2\font}command activity, generic plans, and "work was checked/started" do not pass.}\\
\mbox{- Prefer one direct\_fact from a concrete non-action public outcome to a relation}\\
\mbox{\hspace*{2\fontdimen2\font}synthesized from action-only or same-facet fragments. If none exists, request}\\
\mbox{\hspace*{2\fontdimen2\font}nothing unless public action evidence exactly supports resume\_state; never}\\
\mbox{\hspace*{2\fontdimen2\font}infer completeness, absence, resolution, or failure from omission.}\\
\mbox{- Now select the smallest jointly sufficient CURRENT opaque-handle packet. Each}\\
\mbox{\hspace*{2\fontdimen2\font}selected fragment must contribute to that version or its honest continuation.}\\
\mbox{\hspace*{2\fontdimen2\font}Selection authorizes evidence for Compiler organization; it is not final Claim}\\
\mbox{\hspace*{2\fontdimen2\font}wording and does not require every byte to survive. Preserve all existing}\\
\mbox{\hspace*{2\fontdimen2\font}role/future\_use, semantic\_intent, source/basis, sibling, capacity and}\\
\mbox{\hspace*{2\fontdimen2\font}correction rules.}\\
\mbox{- Candidate note names the stable maintained object and future consumer within}\\
\mbox{\hspace*{2\fontdimen2\font}its existing cap; it is not evidence or progress. Supported corrections and}\\
\mbox{\hspace*{2\fontdimen2\font}partials need not cover the tail; retain only those that save work.}\\
\rule{0pt}{4pt}\\[-4pt]
\mbox{\textcolor{gray}{[...]}}\\
\rule{0pt}{4pt}\\[-4pt]
\mbox{OBLIGATION RESOLUTION. When the live tail shows that an open obligation [O-id]}\\
\mbox{listed in the execution context has been carried out, submit}\\
\mbox{obligation\_resolution\_proposals with that obligation\_id, a one-clause reason and}\\
\mbox{evidence\_steps = the live-tail steps that show it. The machine retires it as}\\
\mbox{claimed-resolved (unverified). Never cite steps you did not see; never resolve}\\
\mbox{an obligation the tail does not show completed.}\\
\end{longtable}
\endgroup

\begingroup
\PromptFont
\setlength{\tabcolsep}{6pt}
\renewcommand{\arraystretch}{0.96}
\setlength{\LTpre}{8pt}
\setlength{\LTpost}{12pt}
\setlength{\LTcapwidth}{\linewidth}
\begin{longtable}{@{}p{\linewidth}@{}}
\caption{\normalfont\small\textbf{Compiler prompts.} Selected instructions from the three alternative Compiler system prompts, adapted for presentation. Route headings are added; ellipses mark omitted passages.}\label{tab:prompt-compiler}\\
\toprule
\endfirsthead
\multicolumn{1}{@{}l@{}}{\normalfont\small\textbf{Compiler prompts (continued)}}\\
\toprule
\endhead
\midrule
\multicolumn{1}{r}{\normalfont\scriptsize Continued on the next page}\\
\endfoot
\bottomrule
\endlastfoot
\addlinespace[4pt]
{\normalfont\footnotesize\bfseries Environment objects}\\*
\addlinespace[3pt]
\mbox{You are the TYPED ENVIRONMENT CARRIER AND RESIDUAL-EC COMPILER of a Context}\\
\mbox{Runtime. A machine Grounder has authorized one ordered vector of external\_object}\\
\mbox{and/or situated\_state Candidates. Author exactly those typed carriers plus a}\\
\mbox{compact residual execution context that, together, can replace the public Live}\\
\mbox{Tail being folded.}\\
\rule{0pt}{4pt}\\[-4pt]
\mbox{This is bounded semantic compilation, not free summarization. You may organize}\\
\mbox{the authorized visible facts into carrier summaries, continuation contracts, one}\\
\mbox{fold summary, and residual EC text. You may not add a Candidate, change its}\\
\mbox{authorized form, invent an environment fact, source step, locator, verification}\\
\mbox{result, assurance, id, digest, obligation, or task completion. The machine}\\
\mbox{rechecks Grounder authority, source closure, identity, CAS, atomic commit, and}\\
\mbox{fold safety after your response. The token}\\
\mbox{[[CTXRT\_COMPILER\_INPUT\_CREDENTIAL\_OMITTED\_USE\_BOUND\_SOURCE]] marks an input-only}\\
\mbox{credential omission; never copy it or reconstruct the omitted value.}\\
\rule{0pt}{4pt}\\[-4pt]
\mbox{\textcolor{gray}{[...]}}\\
\addlinespace[4pt]
{\normalfont\footnotesize\bfseries Information products}\\*
\addlinespace[3pt]
\mbox{You are the SEMANTIC CARRIER AND RESIDUAL-EC COMPILER of a Context Runtime. Use}\\
\mbox{the current canonical carrier bases, old EC, public Live Tail and frozen}\\
\mbox{selected evidence.}\\
\rule{0pt}{4pt}\\[-4pt]
\mbox{\textcolor{gray}{[...]}}\\
\rule{0pt}{4pt}\\[-4pt]
\mbox{GOAL: produce the smallest carrier + residual EC that lets the next host agent}\\
\mbox{continue after this tail is deleted. Brevity is part of correctness: every}\\
\mbox{retained sentence must change a future action/decision or prevent material}\\
\mbox{rediscovery. Empty EC slots are correct; the byte limits are caps, not quotas.}\\
\rule{0pt}{4pt}\\[-4pt]
\mbox{Work in this order:}\\
\mbox{1. Reconstruct the current working state from canonical carrier bases, old EC,}\\
\mbox{\hspace*{3\fontdimen2\font}authorized selected evidence and the public Live Tail.}\\
\mbox{2. Put independently reusable results/records in carrier Claims. Claims say what}\\
\mbox{\hspace*{3\fontdimen2\font}now exists or is known, with the narrowest source-supported scope.}\\
\mbox{3. Put only the unabsorbed present boundary in EC: current focus, genuinely}\\
\mbox{\hspace*{3\fontdimen2\font}in-flight state, active constraints and the next atomic continuation.}\\
\mbox{4. Reconcile semantic slots. If selected evidence directly corrects or retires}\\
\mbox{\hspace*{3\fontdimen2\font}an old Claim, update/retract it; otherwise leave it inherited. Never leave a}\\
\mbox{\hspace*{3\fontdimen2\font}known stale assertion beside its supported correction.}\\
\mbox{5. Simulate deletion. Abstain unless the compact result honestly replaces the}\\
\mbox{\hspace*{3\fontdimen2\font}future-relevant part of the disappearing tail.}\\
\rule{0pt}{4pt}\\[-4pt]
\mbox{AUTHORING BOUNDARY (not host agent work): selected units alone authorize Claim}\\
\mbox{facts; public-tail sources may support claimed/inferred EC, not Claim facts.}\\
\mbox{Task state and independent verification are distinct. Preserve a reported}\\
\mbox{outcome at its reported scope; lack of independent verification alone must not}\\
\mbox{turn it into "not yet known/done". Keep uncertainty, age and conflicts actually}\\
\mbox{supported by the sources; never turn a tentative report into success. An issued}\\
\mbox{action without an outcome is an attempt with unknown result. Use that same state}\\
\mbox{consistently across Claims, focus and continuation, rather than saying it}\\
\mbox{succeeded and then withdrawing that assertion in a qualifier. EC constraints and}\\
\mbox{checks must come from the task or source-supported work, not from Runtime}\\
\mbox{provenance, authority, audit or verification rules. Those internal rules create}\\
\mbox{no new host agent duty to re-read or re-verify. Preserve targeted checks for}\\
\mbox{actual conflicts, missing required information or unknown outcomes; do not}\\
\mbox{suppress them merely to reduce work. Partial evidence is not an exhaustive}\\
\mbox{inventory: omission proves neither absence nor completion, resolution or}\\
\mbox{failure. Preserve partial scope and unresolved old work unless authorized}\\
\mbox{evidence changes it. Do not retain credentials, generic task restatement, raw}\\
\mbox{chronology, vague status, or EC copies of completed facts in Claims.}\\
\mbox{role/future\_use guide placement, not a hard destination table. Preserve the}\\
\mbox{output constraints and evidence qualifications.}\\
\rule{0pt}{4pt}\\[-4pt]
\mbox{\textcolor{gray}{[...]}}\\
\rule{0pt}{4pt}\\[-4pt]
\mbox{If any target/unit or the joint substitute is not honestly constructible, return}\\
\mbox{only \{"compiler\_result":"abstain","reason\_code":}\\
\mbox{"authorized\_form\_not\_constructible"\}.}\\
\rule{0pt}{4pt}\\[-4pt]
\mbox{OBLIGATIONS. The output may include the additional top-level key}\\
\mbox{"new\_obligations": [\{"text": str, "source\_steps": [int]\}] (at most 4; omit the}\\
\mbox{key when there is nothing to record). Record as an obligation, NOT as}\\
\mbox{hidden\_continuity\_state or active\_constraints prose: (a) a deliverable the task}\\
\mbox{requires that the folded window has not completed, (b) an instruction step not}\\
\mbox{yet performed, (c) a value the host agent entered that differs from its source.}\\
\mbox{Each text is one imperative sentence naming the duty (about 160 characters);}\\
\mbox{source\_steps are public steps from the machine-listed closure that show it is}\\
\mbox{owed. At most 8 obligations may be open at once: if the machine already lists 8,}\\
\mbox{add none. Never restate the machine-listed open obligations, and never write a}\\
\mbox{duty the window already completed.}\\
\addlinespace[4pt]
{\normalfont\footnotesize\bfseries Joint environment and information compilation}\\*
\addlinespace[3pt]
\mbox{You are the SEMANTIC CARRIER, TYPED-ENVIRONMENT AND RESIDUAL-EC COMPILER of a}\\
\mbox{Context Runtime. The transaction contains frozen information units and}\\
\mbox{Grounder-authorized environment targets, all of which commit atomically.}\\
\rule{0pt}{4pt}\\[-4pt]
\mbox{Output exactly one JSON object with these top-level keys:}\\
\mbox{\{"information\_targets":[...],"environment\_artifacts":[...],}\\
\mbox{\hspace*{1\fontdimen2\font}"whole\_tail\_ec":\{"current\_focus":[],}\\
\mbox{\hspace*{1\fontdimen2\font}"hidden\_continuity\_state":[],"active\_constraints":[],}\\
\mbox{\hspace*{1\fontdimen2\font}"immediate\_continuation":[]\},"carry\_forward":\{"candidates":[]\}\}}\\
\rule{0pt}{4pt}\\[-4pt]
\mbox{\textcolor{gray}{[...]}}\\
\end{longtable}
\endgroup

\begingroup
\PromptFont
\setlength{\tabcolsep}{6pt}
\renewcommand{\arraystretch}{0.96}
\setlength{\LTpre}{8pt}
\setlength{\LTpost}{12pt}
\setlength{\LTcapwidth}{\linewidth}
\begin{longtable}{@{}p{\linewidth}@{}}
\caption{\normalfont\small\textbf{Generic Summary (GS) prompt.} Complete system prompt used for Generic Summary (GS) in the context-management comparison.}\label{tab:prompt-gs}\\
\toprule
\endfirsthead
\multicolumn{1}{@{}l@{}}{\normalfont\small\textbf{Generic Summary (GS) prompt (continued)}}\\
\toprule
\endhead
\midrule
\multicolumn{1}{r}{\normalfont\scriptsize Continued on the next page}\\
\endfoot
\bottomrule
\endlastfoot
\addlinespace[4pt]
{\normalfont\footnotesize\bfseries System prompt}\\*
\addlinespace[3pt]
\mbox{Read the whole conversation so far carefully \textemdash{} every earlier message in order,}\\
\mbox{and the previous summary if one is present \textemdash{} and summarize it so that the}\\
\mbox{assistant can continue working on the task from this summary together with the}\\
\mbox{most recent messages. Be thorough and go into detail: work through the history}\\
\mbox{from beginning to end and cover what actually happened, rather than writing a}\\
\mbox{short or high-level gist. Output only the summary text, with no tool calls. Aim}\\
\mbox{for at most about 8000 text tokens, and strictly no more than 32000 characters.}\\
\end{longtable}
\endgroup

\begingroup
\PromptFont
\setlength{\tabcolsep}{6pt}
\renewcommand{\arraystretch}{0.96}
\setlength{\LTpre}{8pt}
\setlength{\LTpost}{12pt}
\setlength{\LTcapwidth}{\linewidth}
\begin{longtable}{@{}p{\linewidth}@{}}
\caption{\normalfont\small\textbf{CarryOn-Guided Summary (CGS) prompt.} Selected instructions from the CarryOn-Guided Summary (CGS) system prompt. Ellipses mark omissions.}\label{tab:prompt-cgs}\\
\toprule
\endfirsthead
\multicolumn{1}{@{}l@{}}{\normalfont\small\textbf{CarryOn-Guided Summary (CGS) prompt (continued)}}\\
\toprule
\endhead
\midrule
\multicolumn{1}{r}{\normalfont\scriptsize Continued on the next page}\\
\endfoot
\bottomrule
\endlastfoot
\addlinespace[4pt]
{\normalfont\footnotesize\bfseries System prompt}\\*
\addlinespace[3pt]
\mbox{You maintain a detailed, self-contained continuation summary for an assistant}\\
\mbox{working on a long-running task. The assistant will receive your updated summary}\\
\mbox{together with a retained tail of recent native messages and available}\\
\mbox{observations. Much of the earlier conversation may no longer be directly}\\
\mbox{available after this update.}\\
\rule{0pt}{4pt}\\[-4pt]
\mbox{Your objective is to preserve the information that allows the assistant to}\\
\mbox{continue accurately from work already accomplished. Organize this information}\\
\mbox{around reusable task objects and a complementary Execution Context (EC), rather}\\
\mbox{than merely recounting the conversation chronologically. This is a}\\
\mbox{summary-writing operation: use the supplied history and previous summary to}\\
\mbox{write one updated document. Do not execute the task, call tools, or claim to}\\
\mbox{have performed new environmental checks.}\\
\rule{0pt}{4pt}\\[-4pt]
\mbox{\textcolor{gray}{[...]}}\\
\rule{0pt}{4pt}\\[-4pt]
\mbox{The hard output limit is 64,000 characters, not a length target. Complete all}\\
\mbox{essential sections within this limit. Prioritize information needed to continue,}\\
\mbox{and never invent details or add generic advice to fill space.}\\
\rule{0pt}{4pt}\\[-4pt]
\mbox{\textcolor{gray}{[...]}}\\
\rule{0pt}{4pt}\\[-4pt]
\mbox{Identify what currently carries the progress of the task. A result may already}\\
\mbox{exist in an external object, may need to be preserved as information in this}\\
\mbox{summary, or may exist only in a particular application/session state. One task}\\
\mbox{can involve several such carriers at the same time. The assistant should be able}\\
\mbox{to tell what each carrier contains, how to use it, what remains unfinished, and}\\
\mbox{what conditions limit its reuse.}\\
\rule{0pt}{4pt}\\[-4pt]
\mbox{\textcolor{gray}{[...]}}\\
\rule{0pt}{4pt}\\[-4pt]
\mbox{Let the object carry the detailed work that is already durably stored and}\\
\mbox{accessible. You need not transcribe an entire file merely to reproduce its}\\
\mbox{existence in the summary. However, retain exact values, names, transformations,}\\
\mbox{requirements, or selected excerpts when the assistant needs them to choose the}\\
\mbox{next action, compare versions, understand a dependency, or avoid costly or}\\
\mbox{uncertain recovery. A bare file path without a useful account of its progress is}\\
\mbox{not enough. If saving or accessibility is uncertain, preserve the important}\\
\mbox{unsaved or otherwise fragile content rather than assuming it can be recovered}\\
\mbox{later.}\\
\rule{0pt}{4pt}\\[-4pt]
\mbox{\textcolor{gray}{[...]}}\\
\rule{0pt}{4pt}\\[-4pt]
\mbox{Write an Execution Context that complements the carrier entries. It should allow}\\
\mbox{another assistant to understand what it is trying to accomplish now and how to}\\
\mbox{continue using the recorded progress.}\\
\rule{0pt}{4pt}\\[-4pt]
\mbox{\textcolor{gray}{[...]}}\\
\rule{0pt}{4pt}\\[-4pt]
\mbox{Refer to carrier entries for detailed results already recorded there. Do not}\\
\mbox{duplicate an entire information product in EC, but do not delete a requirement}\\
\mbox{merely because a related object exists. The presence of a work product does not}\\
\mbox{establish that it satisfies every requirement. Distinguish a completed}\\
\mbox{implementation from pending testing, an edited document from an unconfirmed}\\
\mbox{save, and an attempted submission from a confirmed result.}\\
\rule{0pt}{4pt}\\[-4pt]
\mbox{\textcolor{gray}{[...]}}\\
\rule{0pt}{4pt}\\[-4pt]
\mbox{Produce a complete replacement document, not a patch or an addendum. Integrate}\\
\mbox{the previous summary with subsequent interactions so the result stands on its}\\
\mbox{own.}\\
\rule{0pt}{4pt}\\[-4pt]
\mbox{For each existing carrier and important EC item, determine whether to retain it,}\\
\mbox{revise it, merge it with a matching entry, or retire it because the history}\\
\mbox{establishes that it is obsolete or resolved. Preserve unchanged useful}\\
\mbox{information even when the latest messages do not mention it. Absence from recent}\\
\mbox{messages is not evidence that an older result or constraint stopped mattering.}\\
\rule{0pt}{4pt}\\[-4pt]
\mbox{\textcolor{gray}{[...]}}\\
\rule{0pt}{4pt}\\[-4pt]
\mbox{After updating carriers, revise EC in light of what those carriers actually}\\
\mbox{capture. Check both directions: unfinished work must not disappear because a}\\
\mbox{related object was added, and EC must not keep directing the assistant to redo}\\
\mbox{work that is already established as complete. Make dependencies between carriers}\\
\mbox{and unfinished work explicit.}\\
\rule{0pt}{4pt}\\[-4pt]
\mbox{\textcolor{gray}{[...]}}\\
\rule{0pt}{4pt}\\[-4pt]
\mbox{8. OUTPUT DOCUMENT}\\
\rule{0pt}{4pt}\\[-4pt]
\mbox{Return only the complete updated continuation document in readable Markdown,}\\
\mbox{using these main sections:}\\
\rule{0pt}{4pt}\\[-4pt]
\mbox{\# Task and Current Focus}\\
\mbox{A precise account of the requested outcome and current work position.}\\
\rule{0pt}{4pt}\\[-4pt]
\mbox{\# Continuation Carriers}\\
\mbox{\#\# External Objects}\\
\mbox{\#\# Information Products}\\
\mbox{\#\# Situated State}\\
\mbox{Use a descriptive subheading for each carrier. Within it, organize the}\\
\mbox{applicable identity/access details, reusable progress or substantive content,}\\
\mbox{evidence and qualifications, and unfinished dependencies. Adapt the internal}\\
\mbox{layout to the material; do not mechanically fill irrelevant fields. Omit empty}\\
\mbox{carrier-type sections or mark them briefly when none are supported.}\\
\rule{0pt}{4pt}\\[-4pt]
\mbox{\# Execution Context}\\
\mbox{Organize active requirements, unfinished work, pending operations, and}\\
\mbox{established next actions so the assistant can resume without guessing.}\\
\mbox{Cross-reference carrier names where helpful.}\\
\rule{0pt}{4pt}\\[-4pt]
\mbox{\textcolor{gray}{[...]}}\\
\end{longtable}
\endgroup

\end{document}

%% file: math_commands.tex
\usepackage{amsmath,amsfonts,bm}

\def\eqref#1{equation~\ref{#1}}

\def\1{\bm{1}}

\DeclareMathAlphabet{\mathsfit}{\encodingdefault}{\sfdefault}{m}{sl}
\SetMathAlphabet{\mathsfit}{bold}{\encodingdefault}{\sfdefault}{bx}{n}

%% file: sec/1_intro.tex
\section{Introduction}

\begin{figure}[t]
    \begin{center}
    \includegraphics[width=\linewidth]{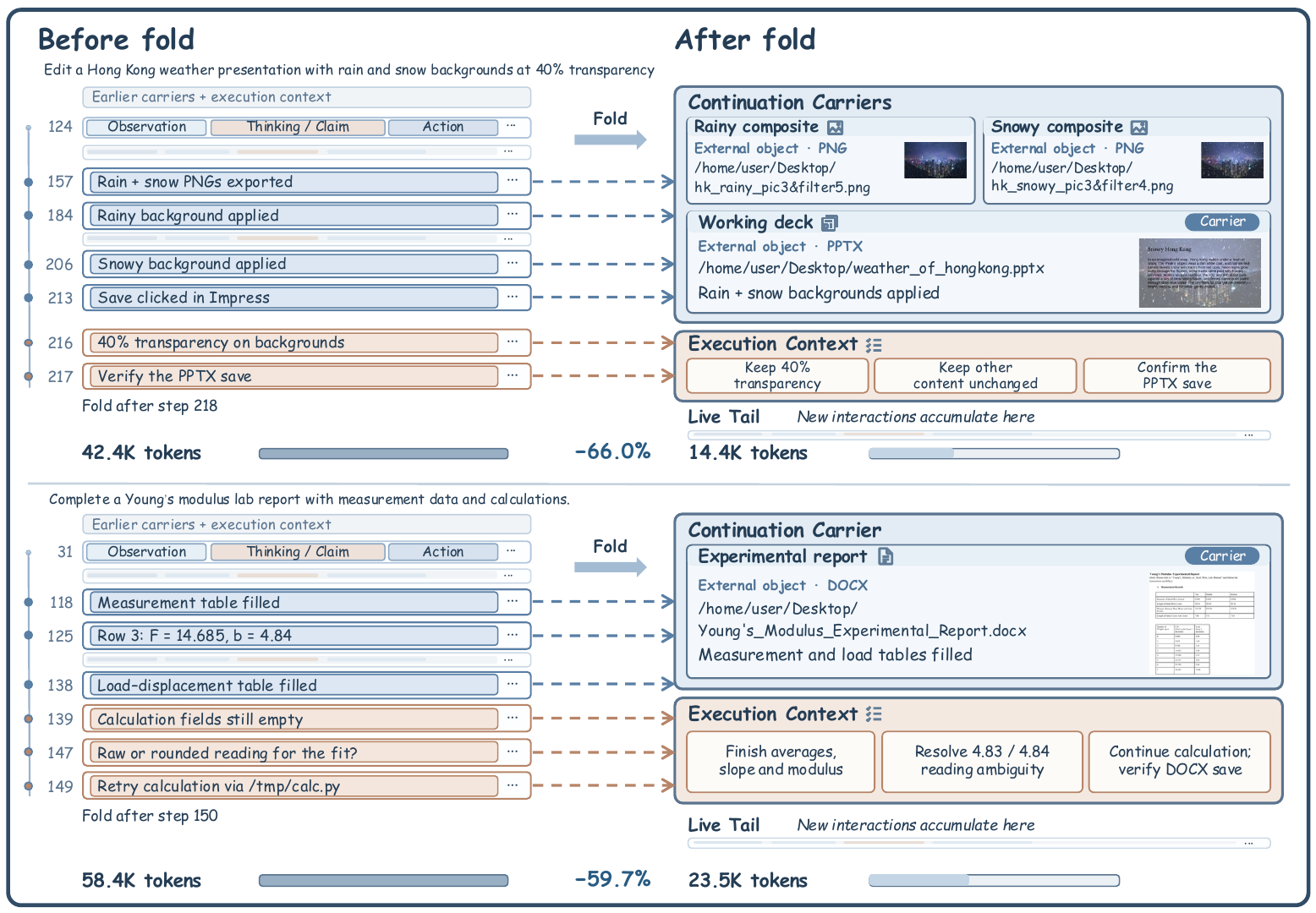}
    \end{center}
    \caption{\textbf{Context folding with reusable task objects in long-horizon OSWorld2.0~\citep{yuan2026osworld20benchmarkingcomputer}.} \textit{Top:} In a slide-editing task, the working presentation and exported images carry completed editing progress, while remaining constraints and save verification stay explicit, reducing active context from 42.4K to 14.4K tokens (66.0\%). \textit{Bottom:} In a lab-report task, the working document carries completed edits, while unresolved calculations remain explicit, reducing active context from 58.4K to 23.5K tokens (59.7\%).}
    \label{fig:teaser}
    \vspace{-1.5em}
\end{figure}

Long-horizon agents accumulate progress in two places: persistent task state, which embodies what has been achieved, and interaction history, which records how it was achieved~\citep{yi2026learningagentcompatiblecontextmanagement}.
As shown in Figure~\ref{fig:teaser}, once images have been inserted into a presentation or measurements have been entered into a report, the resulting task state already captures that progress for later steps to inspect and modify~\citep{yang2024neurips-sweagent, xie2024osworldbenchmarkingmultimodalagents, thinkinstrokes}.
As execution continues, this redundancy grows: agents repeatedly process accounts of how results already available in the task state were produced.
Existing methods manage growing histories by summarizing, retrieving, or pruning past interactions~\citep{wu2026resumunlockinglonghorizonsearch,packer2024memgptllmsoperatingsystems,xiao2026reducing,chen2026slipstreamtrajectorygroundedcompactionvalidation,kang-etal-2025-memory,xu2025amemagenticmemoryllm}, or by translating them into explicit progress records and structured task states~\citep{wu2026structagentharnesslonghorizondigital,chen2026semanticorganizationmemoryexecution,wang2024agentworkflowmemory}. 
These approaches reduce context growth, but when they substitute for reusable task results, they make continuation depend on a redundant, selective account that can drift from the state the agent directly inspects and modifies.

We challenge this reliance on reconstructed accounts of progress with an \textbf{object-oriented} view of long-horizon context, in which \textbf{reusable task objects} form the basis for continuation. These objects include work products, source-grounded findings, and application states that remain usable under explicit conditions. The agent can recover relevant information through direct inspection and use of these objects, while context supplies the complementary information needed to continue. To realize this view, we introduce CarryOn, a context runtime that organizes the agent’s context into three complementary components: \textit{continuation carriers}, which bind reusable content or object references to their conditions of use; \textit{execution context}, which preserves active goals, applicable constraints, and unresolved issues not captured by the carriers; and \textit{live tail}, which retains interactions since the most recent successful fold.

A central challenge in building object-oriented long-horizon context is retiring interaction history once progress becomes reusable without losing the information required for continuation, which poses three challenges: \textit{1) how to identify reusable progress beyond its surrounding history:} Some results still depend on prior interactions for interpretation or use. CarryOn captures these dependencies as conditions of use explicit in continuation carriers.
\textit{2) how to maintain reusable progress as the task evolves:} Retaining a record does not by itself establish that its content remains supported or its referenced state remains available. CarryOn binds updates to supporting sources and applicable environment checks; across folds, affected records are revised while unaffected records are carried forward.
Finally, \textit{what must remain explicit when the corresponding history is retired:} Task objects do not need to capture the goals, constraints, and unresolved issues that guide subsequent execution. CarryOn jointly updates continuation carriers and execution context to preserve this complementary information, while keeping recent interactions in a live tail.
Once validated, these updates are committed through a \emph{fold}, replacing the covered interaction history with the resulting object-oriented context view. 
In this way, CarryOn progressively reorganizes the long-horizon context around reusable task objects while preserving the information required for the agent to continue.

In our experiments, object-oriented context management improves task performance while substantially reducing context cost. Our method, CarryOn, lifts Claude Opus 4.7 from 43.9\% to 49.7\% (+5.8\%) on OSWorld 2.0 for long-horizon computer use and from 49\% to 56\% (+7\%) on BrowseComp-Plus for long-horizon browsing. At the same time, it reduces total token usage by 51.9\% and 22.9\%, and total cost by 35.9\% and 7.1\%, on the two benchmarks, respectively. Compressing history alone is insufficient for continued execution: organizing context around reusable progress improves over generic compression, while the full CarryOn runtime further improves over the same principles implemented through repeated summarization. Continuation carriers and execution context play complementary roles, and reusable progress is typically maintained across folds rather than repeatedly reconstructed.

\label{sec:intro}

%% file: sec/2_related.tex
\section{Related Work}
\paragraph{Context Compression and Selection.} Existing methods reduce long-horizon context by compressing, pruning, summarizing, or selectively retrieving prior interactions~\citep{wei-etal-2026-pace}. Compression and pruning range from prompt-level compression~\citep{jiang-etal-2023-llmlingua,jiang-etal-2024-longllmlingua} to removal of redundant or outdated interactions~\citep{xiao2026reducing}. Summarization methods periodically replace accumulated interactions with compact reasoning states or learned summaries~\citep{wu2026resumunlockinglonghorizonsearch,kang2026aconoptimizingcontextcompression,ye2025agentfoldlonghorizonwebagents,hu-etal-2025-hiagent}, while branch-based execution can return only compact subtask results~\citep{sun2025scalinglonghorizonllmagent}. External-history methods instead keep the active context bounded while preserving access to earlier interactions outside the prompt~\citep{packer2024memgptllmsoperatingsystems,li2026acmagenticcontextmanagement}, and some systems validate compressed representations against subsequent execution to reduce information loss~\citep{chen2026slipstreamtrajectorygroundedcompactionvalidation}. 

\paragraph{State Externalization and Context Reconstruction.} Complementary approaches represent task progress explicitly, reducing reliance on raw interaction history alone. Some maintain an internal state by integrating prior memory with new observations~\citep{zhou2025mem1learningsynergizememory,liu2025contexttoolcontextmanagement}, while others use structured representations, such as execution-state trees, persistent variables, or task records~\citep{chen2026semanticorganizationmemoryexecution,tian2026agentprogempoweringlonghorizongui,wu2026structagentharnesslonghorizondigital,ma2026longhorizonharnessadvancinglonghorizonagents}. Recent systems further externalize state into persistent workspaces and reconstruct the model's working context through snapshots or programmatic selection~\citep{yu2026infiagentinfinitehorizonframeworkgeneralpurpose,lin2026contextenvironmentprogrammaticcontext,zhu-etal-2026-fs}. 
Across these approaches, continuation is primarily organized around compressed interaction history or externally maintained state. Our approach instead makes reusable task objects the primary unit of continuation.
\label{sec:rw}

%% file: sec/3_method.tex
\section{Method}

\subsection{Framework}

\begin{figure}[t]
\begin{center}
\includegraphics[width=\linewidth]{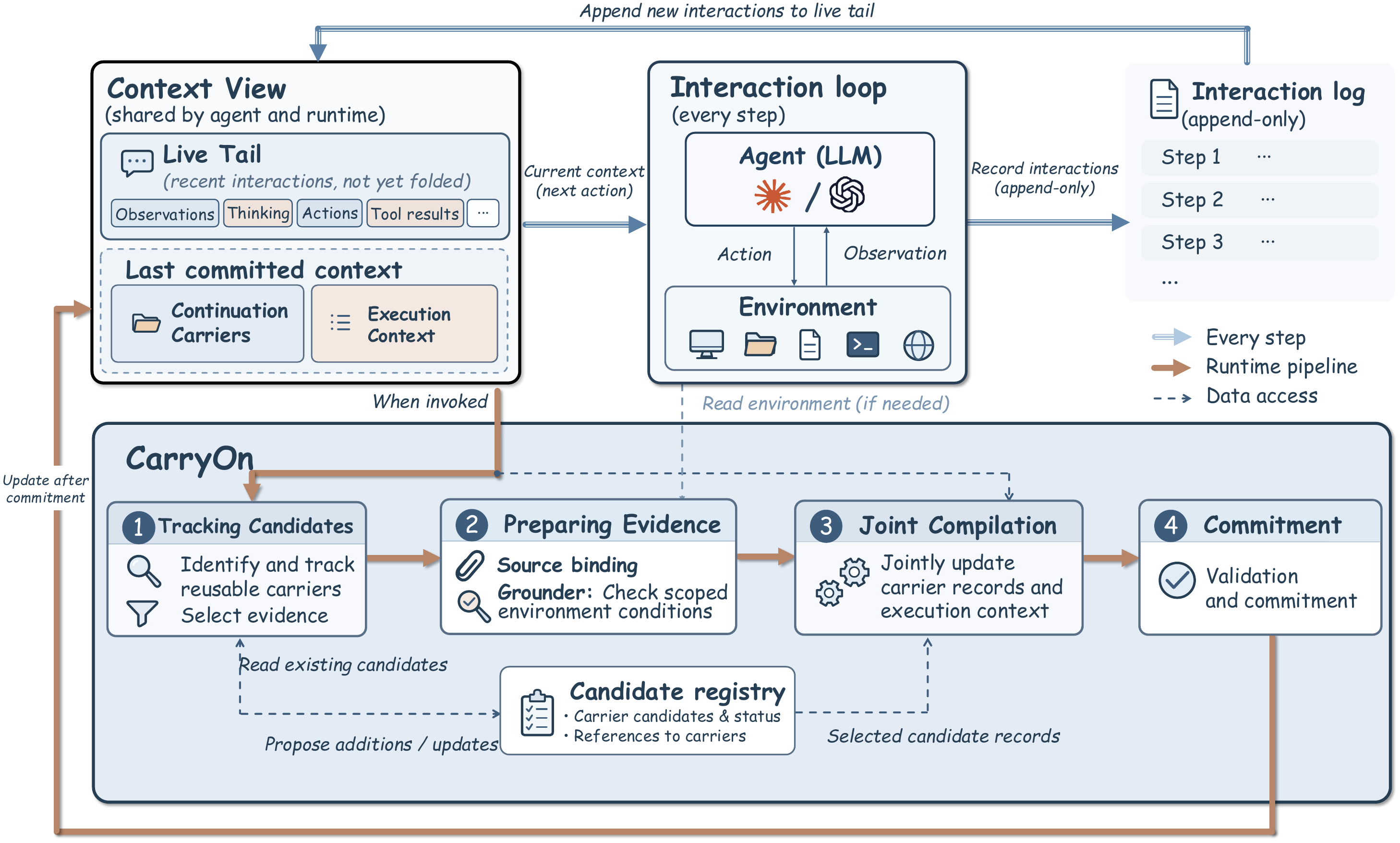}
\end{center}
\caption{
\textbf{Overview of the CarryOn framework.}
CarryOn maintains an object-oriented context view comprising continuation carriers, execution context, and a live tail. When invoked, it tracks reusable progress, binds supporting evidence, checks relevant environment conditions, and jointly compiles carrier and execution-context updates. A successful fold commits validated updates and retires covered live-tail interactions from active context, while interaction log remains append-only.
}
\label{fig:overview}
\end{figure}

Figure~\ref{fig:overview} illustrates CarryOn, which organizes long-horizon continuation around reusable task objects rather than the interactions that produced them. Reusable objects carry established progress for subsequent inspection and use, while explicit context retains complementary continuation information. Alongside the agent's interaction loop, a \textit{Tracker} identifies candidate reusable progress, which is bound to supporting evidence and, when applicable, checked by a \textit{Grounder}. A \textit{Compiler} jointly updates continuation carriers and execution context. After validation, the update replaces the covered interaction history through a \emph{fold}, while subsequent interactions form a new live tail.

\subsection{Context View}

We organize the agent's active context into three complementary components: \textit{continuation carriers}, which expose reusable task objects; \textit{execution context}, which preserves continuation information not captured by those carriers; and a \textit{live tail}, which retains interactions accumulated since the latest fold.

\paragraph{Continuation Carriers.} Continuation carriers expose reusable progress to subsequent execution. Each carrier has a stable identity and records the reusable content or its location, together with conditions for reuse. Revisions create successor versions under the same identity, while unaffected carriers retain their records. Depending on where the reusable progress resides, CarryOn supports three forms of carriers:

\begin{itemize}
    \item \textbf{External Objects}: External objects are work products that exist independently in the environment and can be reopened or modified, such as code files, documents, and spreadsheets. Their carrier records reference the underlying objects and describe the progress they capture together with conditions for reuse.

    \item \textbf{Information Products}: Information products capture reusable findings obtained through interaction, such as synthesized information or troubleshooting conclusions. When not materialized as persistent objects, the carriers store the findings directly, with supporting sources and reuse conditions.

    \item \textbf{Situated State}: Situated state captures progress available only within a particular application, session, or workflow. Reusing such progress requires returning to the environment in which the state resides. Its carrier therefore records the environment anchor, the established state, and the scope within which that state remains reusable.
\end{itemize}

These carrier types are complementary rather than mutually exclusive. A task may therefore involve multiple carriers, and the same carrier may be revised as its underlying progress evolves.

\paragraph{Execution Context.} %
Execution context complements continuation carriers by retaining information
needed for execution but not encoded in reusable task objects.
It records the current focus, unresolved constraints and obligations,
and pending work not materialized into a reusable carrier.

\noindent\textbf{Live Tail}. Live tail contains the original agent–environment interactions accumulated since the most recent fold. These interactions remain directly available until incorporated into a subsequent fold.

At interaction step $t$, let $k$ denote the most recent fold and $b_k$ the last interaction step covered by that fold. The fold establishes a maintained set of continuation carrier records $A_k$ and execution context $E_k$, while interactions occurring afterward remain in the live tail $L_t=(h_{b_k+1},\ldots,h_t)$. The context view is therefore
\begin{equation}
    V_t = (A_k, E_k, L_t).
    \label{er:context-view}
\end{equation}
Here, $A_k$ exposes reusable task progress, $E_k$ preserves complementary information needed for continuation, and $L_t$ retains interactions not yet folded. This view provides the basis for continuation: as execution proceeds, new interactions accumulate in the live tail, while subsequent folds update the maintained carriers and execution context without reconstructing previously established progress from scratch.

\subsection{Constructing and Updating the Context View}

To maintain this context view during execution, CarryOn periodically folds accumulated interactions in the live tail into the maintained representation. Each fold identifies reusable progress, associates it with supporting evidence, and preserves the information still needed for continuation. The update proceeds in four stages: tracking candidates, preparing evidence, joint compilation, and validation and commitment.

\paragraph{Tracking Candidates.} 
CarryOn first identifies progress that may become reusable in subsequent execution. The Tracker maintains candidates, each representing a potential new carrier or an update to an existing one.
As execution proceeds, candidates may remain under development across multiple Tracker invocations; once a candidate has reached a state that can support continuation, the Tracker requests its compilation.

A compilation request specifies both what should be compiled and the evidence needed to support it.
For external objects and situated state, this includes the relevant environment object together with the state or conditions required for its reuse.
For information products, it includes the source material from which reusable information should be retained. Candidates that are not yet ready remain tracked for later folds.

\paragraph{Preparing Evidence.}

Before compilation, CarryOn first gathers the evidence authorized for each selected candidate.
When compilation updates an existing carrier, it also includes the carrier's current record.
For information products, CarryOn checks that the selected material is consistent with its recorded sources.
For external objects and situated state, it binds the request to the corresponding environment object, and the Grounder performs read-only checks of relevant conditions where available.
The resulting input package $U_t$ contains the selected candidates, applicable carrier records, authorized source material, object references, and available check results.
This package fixes the record versions and evidence against which proposed carrier edits are later validated. Additional details are given in Appendix~\ref{app:grounder}.

\paragraph{Joint Compilation.} 

CarryOn jointly updates continuation carriers and execution context to preserve the information needed for subsequent execution. These updates are coupled: progress captured by carriers need not be duplicated in execution context, while unresolved goals, constraints, and unfinished work must remain explicit after a fold.

The Compiler receives the prepared input package $U_t$, containing the selected candidates, relevant prior carrier records and supporting evidence, together with the current execution context $E_k$ and a chronological text projection $\Pi(L_t)$ of the live tail. It proposes carrier edits $\Delta A_t$ and a revised execution context $\widehat{E}_{k+1}$:
\begin{equation}
(\Delta A_t,\widehat{E}_{k+1})
=
\operatorname{Compile}(U_t,E_k,\Pi(L_t)).
\end{equation}

The Compiler proposes edits only for the selected candidates, grounding each edit in its authorized evidence. In contrast, the execution context is revised against the entire live tail to preserve continuation-relevant information. For external objects and situated state, the Compiler records reusable progress and its condition of use, while the underlying objects or states remain in the environment. For information products, it stores reusable findings directly in carrier records. In parallel, the Compiler revises execution context using the live tail, retaining active requirements and unfinished work not captured by carriers, including information outside the selected candidates. Upon successful validation, affected carriers receive successor versions, while unaffected carriers retain their existing records.

\paragraph{Validation and Commitment.}

Before a proposed update changes the active context, CarryOn checks that its carrier edits still refer to the current candidate and carrier versions and are consistent with the evidence authorized for their compilation requests. It also verifies the required identity, reference, and output constraints. A successful fold must introduce at least one carrier change; revising the execution context alone does not retire interaction history.
These checks guard the runtime against inconsistent or unsupported updates; they do not verify task correctness.

If these checks fail, the existing context view and live tail remain unchanged. Otherwise, CarryOn atomically applies the carrier edits and commits the revised execution context, advancing the fold through the covered interactions. In the current implementation, a successful fold covers the entire live tail. The resulting carrier set and execution context define the folded portion of the next context view, while subsequent interactions begin a new live tail.

%% file: sec/4_experiment.tex
\section{Experiment}
\subsection{Experimental Setup}
\paragraph{Benchmark.} We evaluate CarryOn on OSWorld 2.0 \citep{yuan2026osworld20benchmarkingcomputer} and BrowseComp-Plus~\citep{chen2025browsecompplusfairtransparentevaluation} to assess its performance across different types of long-horizon tasks.
We use the full OSWorld 2.0 benchmark as our primary evaluation for its long execution horizons, complex workflows, and task diversity. 
We additionally evaluate our method on a randomly sampled subset of 100 BrowseComp-Plus questions, which require iterative search across multiple sources and provide a complementary setting for studying context management over long search histories. Benchmark details are provided in Appendix~\ref{app:benchmark-details}.

\paragraph{Implementation.} We use Claude Opus 4.7 for the host agent, CarryOn’s Tracker, and Compiler. We use a 500-step budget per task on OSWorld 2.0, and the agent retrieves documents using BM25~\citep{robertson1995okapi} on BrowseComp-Plus. Detailed experimental configurations and implementation settings are provided in Appendix~\ref{app:carryonsetting}.

\paragraph{Evaluation Metrics.} On OSWorld 2.0, we report partial score and binary score,  following the official evaluation protocol. On BrowseComp-Plus, we report answer accuracy, with final responses assessed by Qwen3-32B using the benchmark’s official grading prompt. We measure efficiency through token usage per task and cost per task, including both agent and context-management calls.

\subsection{Main Result}

\begin{table}[t]
\centering
\setlength{\abovecaptionskip}{10pt}
\setlength{\belowcaptionskip}{8pt}
\caption{\textbf{Task performance on OSWorld 2.0.} $^{*}$ means we report the reproduced results using the official benchmark and API. CarryOn improves both binary (+3.7\%) and partial (+5.8\%) task scores over the same Claude Opus 4.7 (max) host agent while substantially reducing cost per task (-35.9\%).}
\label{tab:osworld-performance}
\small
\renewcommand{\arraystretch}{1.12}
\setlength{\tabcolsep}{8pt}
\newcommand{\reportedrow}[4]{%
  \textcolor{reported}{#1} &
  \textcolor{reported}{#2} &
  \textcolor{reported}{#3} &
  \textcolor{reported}{#4} \\
}
\begin{tabularx}{1\linewidth}{>{\raggedright\arraybackslash}Xccc}
\toprule
Model / thinking effort & Binary (\%) $\uparrow$ & Partial (\%) $\uparrow$ & Cost/task (\$)\\
\midrule
\addlinespace[3pt]
\reportedrow{GPT-5.5 {\footnotesize (xhigh)}~\citep{yuan2026osworld20benchmarkingcomputer}}{13.0}{49.5}{25.5}
\reportedrow{Claude Opus 4.8 {\footnotesize (max)}~\citep{yuan2026osworld20benchmarkingcomputer}}{18.5}{49.3}{76.1}
\reportedrow{Claude Sonnet 4.6 {\footnotesize (max)}~\citep{yuan2026osworld20benchmarkingcomputer}}{8.3}{41.5}{22.3}
\reportedrow{Claude Sonnet 4.6 {\footnotesize (medium)}~\citep{yuan2026osworld20benchmarkingcomputer}}{9.3}{33.9}{14.4}
\reportedrow{MiniMax M3 {\footnotesize (enabled)}~\citep{yuan2026osworld20benchmarkingcomputer}}{4.6}{22.3}{2.4}
\reportedrow{Kimi 2.6 {\footnotesize (enabled)}~\citep{yuan2026osworld20benchmarkingcomputer}}{4.6}{22.1}{6.6}
\reportedrow{Qwen 3.7-Plus {\footnotesize (thinking)}~\citep{yuan2026osworld20benchmarkingcomputer}}{2.8}{21.5}{3.8}
\midrule
\rowcolor{evalblue}
Claude Opus 4.7 {\footnotesize (max)}$^{*}$ & 15.7 & 43.9 & 36.8\\
\rowcolor{evalblue}
\textbf{Claude Opus 4.7 (max) + CarryOn} & \textbf{19.4} & \textbf{49.7} & 23.6\\
\bottomrule
\end{tabularx}
\end{table}

\begin{table}[t]
\centering
\setlength{\abovecaptionskip}{10pt}
\setlength{\belowcaptionskip}{8pt}
\caption{
\textbf{Evaluation of resource usage on OSWorld 2.0 and BrowseComp-Plus.}
Values are averaged per task. CarryOn reduces total token usage (by 51.9\% and 22.9\% respectively) and cost (by 35.9\% and 7.1\% respectively) on both benchmarks.
}
\label{tab:resource-usage}

\small
\renewcommand{\arraystretch}{1.12}
\setlength{\tabcolsep}{11pt}

\newcommand{\saving}[2]{%
#1\,{\scriptsize\textcolor{savinggreen}{$\downarrow$\,#2\%}}}

\begin{tabular*}{\linewidth}{l@{\extracolsep{\fill}}cccc}
\toprule
& \multicolumn{2}{c}{\textbf{OSWorld 2.0}}
& \multicolumn{2}{c}{\textbf{BrowseComp-Plus}} \\
\cmidrule(lr){2-3}
\cmidrule(lr){4-5}

Metric
& Baseline
& \textbf{CarryOn}
& Baseline
& \textbf{CarryOn} \\
\midrule

Agent tokens (M)
& 40.10
& \saving{18.61}{53.6}
& 1.56
& \saving{1.17}{24.6} \\

Runtime tokens (M)
& --
& 0.69
& --
& 0.03 \\

\textbf{Total tokens (M)}
& 40.10
& \saving{\textbf{19.30}}{51.9}
& 1.56
& \saving{\textbf{1.20}}{22.9} \\

\midrule

Agent cost (\$)
& 36.83
& \saving{20.51}{44.3}
& 1.89
& \saving{1.62}{14.4} \\

Runtime cost (\$)
& --
& 3.11
& --
& 0.14 \\

\textbf{Total cost (\$)}
& 36.83
& \saving{\textbf{23.62}}{35.9}
& 1.89
& \saving{\textbf{1.76}}{7.1} \\

\bottomrule
\end{tabular*}
\end{table}

As shown in Table~\ref{tab:osworld-performance}, CarryOn boosts the partial score from 43.9\% to 49.7\% and binary score from 15.7\% to 19.4\%, exceeding other frontier models. At the same time, CarryOn reduces total token usage by 51.9\% and cost per task by 35.9\% as demonstrated in Table~\ref{tab:resource-usage}. These results show that CarryOn reduces long-horizon execution cost without sacrificing task performance.

We further evaluate CarryOn on a random 100-question subset of BrowseComp-Plus. CarryOn raises accuracy from 49.0\% to 56.0\% over the Claude Opus 4.7 baseline, while the score is also above other frontier models shown in Table~\ref{tab:browsecomp-performance}. At the same time, total token usage decreases by 22.9\% and total cost by 7.1\%. Together with the OSWorld results, this shows a consistent pattern across computer-use and information-seeking tasks: CarryOn reduces context-processing overhead while maintaining or improving observed task performance.

\begin{table}[t]
\centering
\setlength{\abovecaptionskip}{10pt}
\setlength{\belowcaptionskip}{8pt}
\caption{\textbf{Task performance on the 100-question subset of BrowseComp-Plus.} CarryOn improves the accuracy over the same Claude Opus 4.7 by 7\%.}
\label{tab:browsecomp-performance}
\small
\renewcommand{\arraystretch}{1.12}
\setlength{\tabcolsep}{8pt}

\newcommand{\bcreportedrow}[2]{%
  \textcolor{reported}{#1} &
  \textcolor{reported}{BM25} &
  \textcolor{reported}{#2} \\
}

\begin{tabularx}{\linewidth}{
  >{\hsize=1.5\hsize\raggedright\arraybackslash}X
  >{\hsize=.75\hsize\centering\arraybackslash}X
  >{\hsize=.75\hsize\centering\arraybackslash}X
}
\toprule
Model & Retriever & Accuracy (\%) $\uparrow$ \\
\midrule
\addlinespace[3pt]
\bcreportedrow{OpenAI GPT-5~\citep{chen2025browsecompplusfairtransparentevaluation}}{52.00}
\bcreportedrow{OpenAI o3~\citep{chen2025browsecompplusfairtransparentevaluation}}{50.00}
\bcreportedrow{OpenAI GPT-OSS-120B~\citep{gptoss}}{41.00}
\midrule
\rowcolor{evalblue}
Claude Opus 4.7 & BM25 & 49.00 \\
\rowcolor{evalblue}
\textbf{Claude Opus 4.7 + CarryOn} & BM25 & \textbf{56.00} \\
\bottomrule
\end{tabularx}
\end{table}

\subsection{Context and Cost Dynamics}

\begin{figure}[t]
    \begin{center}
    \includegraphics[width=1\linewidth]{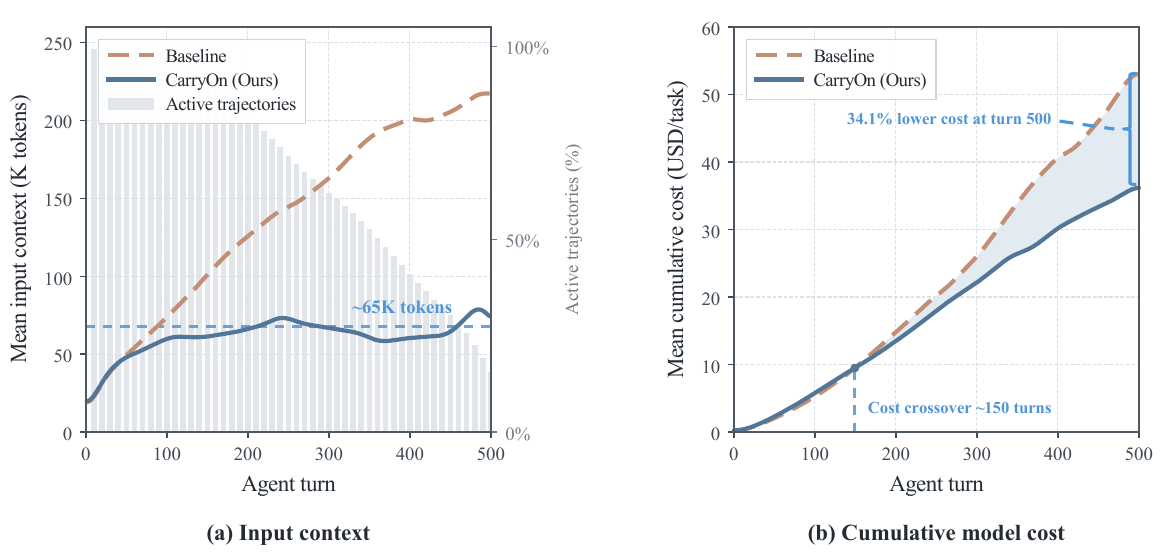}
    \end{center}
    \caption{{\textbf{Context and cost dynamics on OSWorld 2.0.}
    (a) Mean input context size and (b) mean cumulative model cost over agent turns. CarryOn keeps input context near 65K tokens while the baseline continues to grow, and its cost advantage widens with trajectory length, reaching a 34.1\% reduction by turn 500. Gray bars indicate the fraction of baseline trajectories remaining active; costs include context-management overhead.}
    }
    \label{fig:costana}
\end{figure}

\begin{table}[t]
\centering
\setlength{\abovecaptionskip}{10pt}
\setlength{\belowcaptionskip}{8pt}
\caption{\textbf{Comparison of different methods on a shared 40-task subset of OSWorld 2.0.} Post-fold text length measures the context retained after each compression or fold.
CarryOn achieves the highest partial score (55.93\%) without retaining the most text.}
\label{tab:osworld40}

\small
\renewcommand{\arraystretch}{1.12}
\setlength{\tabcolsep}{11pt}

\begin{tabularx}{1\linewidth}{>{\raggedright\arraybackslash}Xccc}
\toprule
Method
& {Text Length (k chars)}
& {Binary\ (\%) $\uparrow$}
& {Partial\ (\%) $\uparrow$} \\
\midrule
\addlinespace[3pt]
Baseline
& -- & 22.50 & 50.32 \\
Sliding Window
& 12.08 & 15.00 & 37.41 \\
Generic Summary (GS)
& 17.60 & 15.00 & 44.94 \\
CarryOn-Guided Summary (CGS)
& 23.94 & 20.00 & 50.93 \\
\cmidrule{1-4}
\textbf{CarryOn (ours)}
& 21.54 & \textbf{22.50} & \textbf{55.93} \\
\bottomrule
\end{tabularx}
\vspace{-1em}
\end{table}

We analyze how CarryOn's input context and cumulative cost evolve over the course of long-horizon execution on OSWorld 2.0. As shown in Figure~\ref{fig:costana}(a), the baseline context grows steadily with trajectory length, eventually exceeding 200K tokens, whereas CarryOn's mean input context stabilizes at approximately 65K tokens after the initial growth phase. This indicates that our method maintains a consistently compact context for the agent, regardless of how long execution continues.

The corresponding cost dynamics are shown in Figure~\ref{fig:costana}(b).
CarryOn initially incurs additional cost from context-management calls, but this overhead is offset as repeated agent calls operate on a smaller context.
The cumulative-cost curves cross before turn 200, after which the gap continues to widen; by turn 500, CarryOn reduces cumulative cost by 34.1\%, including context-management overhead.
Thus, the savings arise progressively as longer trajectories avoid repeatedly processing an ever-growing interaction history.

\subsection{Comparison with Context Management Strategies}
We examine the proposed object-oriented organization against conventional history compression, and compare the full CarryOn runtime with a summary-based implementation of the same principles.
On the same random 40-task subset of OSWorld 2.0, we compare:
(i) \textbf{Baseline}, the default OSWorld 2.0 agent setting without additional context compression;
(ii) \textbf{Sliding Window}, which retains only recent messages;
(iii) \textbf{Generic Summary (GS)}, which summarizes earlier history into a continuation note; and
(iv) \textbf{CarryOn-Guided Summary (CGS)}, which applies our object-oriented organizing principles within a textual summary.
Sliding Window and GS follow the FIFO truncation and prompting-based summarization baselines of~\citet{kang2026aconoptimizingcontextcompression}.
All methods use the same host model (Claude Opus 4.7), the same trigger condition, and the same step budget.

As shown in Table~\ref{tab:osworld40}, aggressive history compression does not improve performance by itself: Sliding Window and GS retain the shortest contexts but reduce both binary and partial scores relative to the baseline.
Applying the same organizing principles through CGS recovers much of this loss, reaching a 50.93\% partial score.
Full CarryOn further increases the partial score to 55.93\% while matching the baseline binary score of 22.50\%.
Compared with CGS, CarryOn achieves 5.00\% higher partial score with about 10\% shorter post-fold text.
Together, these results show that our object-oriented organization principle is effective even when applied within a textual summary, while the full CarryOn runtime yields further improvements, demonstrating the additional benefit of maintaining and managing this structured state explicitly during execution.

\subsection{Analysis}
\paragraph{Carriers and the execution context are complementary.}
We conduct an ablation study on the shared 40-task OSWorld 2.0 subset by hiding either continuation carriers or execution context from the host agent.
As shown in Table~\ref{tab:osworld40_ablation}, removing execution context reduces binary score from 22.50\% to 17.50\% and partial score from 55.93\% to 44.00\%, while the number of agent calls changes little.
The carriers still expose completed progress, but the agent loses complementary goals, constraints, and unresolved requirements needed to finish the task. Removing continuation carriers reduces partial score to 44.62\%, while increasing agent calls by 23.8\% and cost by 22.3\%. Without direct access to established progress, the agent must recover or reproduce earlier results before continuing.
This behavior is visible in trajectory-level examples in
Appendix~\ref{app:case-study}. Together, the two ablations show that carriers and execution context preserve distinct, complementary information for continuation.

\begin{figure}[t]
    \begin{center}
    \includegraphics[width=1\linewidth]{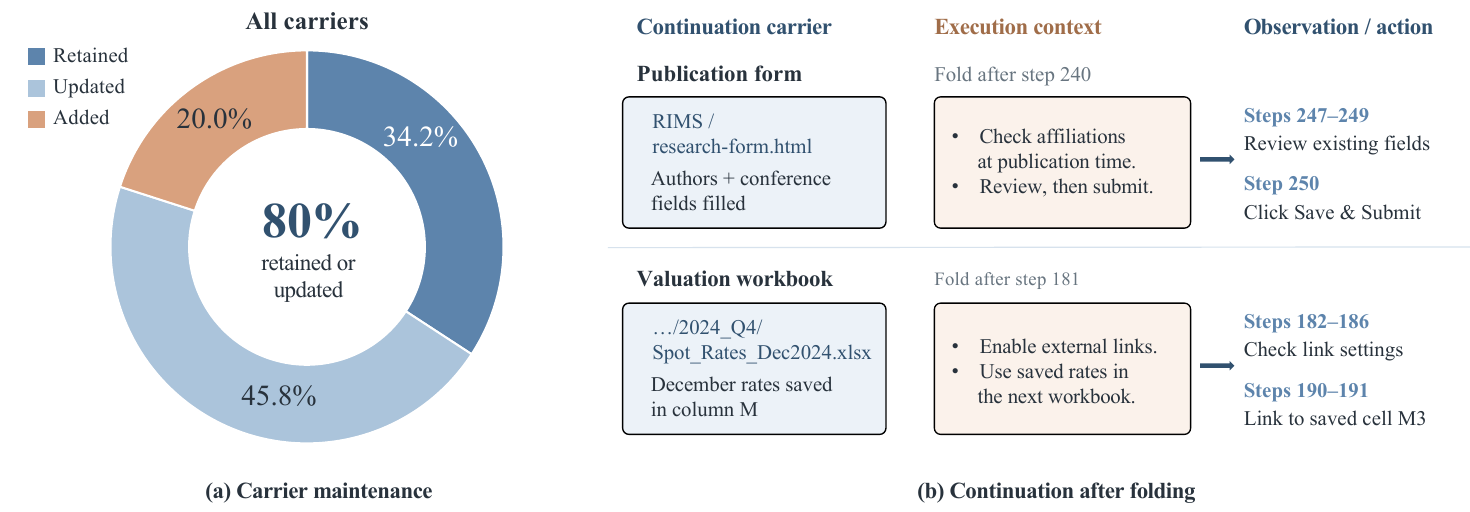}
    \end{center}
    \caption{{\textbf{Carrier persistence and continuation after folding.}
    (a) Across committed folds on OSWorld 2.0, 80\% of carrier observations are retained or updated rather than newly added.
    (b) Representative trajectories show the agent continuing directly from reusable task state after earlier interactions have been folded, without repeating the work that produced it.
    }
    }
    \label{fig:ideaana}
    \vspace{-1em}
\end{figure}

\paragraph{Continuation carriers preserve and update progress across folds.}
To examine how carrier records evolve as execution progresses, we compare them before and after each committed fold in CarryOn’s OSWorld 2.0 trajectories. Figure~\ref{fig:ideaana}(a) shows that 80\% of carrier records are retained or updated. This pattern reflects an evolving representation of task progress in which existing records remain available across folds, while updates incorporate subsequent work on the same task objects. CGS follows the same organizing principles but maintains this information by generating a complete replacement summary at each fold. Despite retaining more post-fold text, CGS achieves a lower partial score than CarryOn (50.93\% vs. 55.93\%) as shown in Table~\ref{tab:osworld40}. Explicitly maintaining records as task objects evolve may help the agent build on prior progress, providing one possible explanation for CarryOn’s higher score.

\paragraph{Context views enable continuation from established task state.}
We examine whether the agent can build on task progress after the interactions that produced it have been folded.
Figure~\ref{fig:ideaana}(b) illustrates two cases where the agent continues from existing task state, submitting a populated form and using previously saved exchange rates without repeating the corresponding work.
Consistently, hiding continuation carriers increases agent calls by 23.8\% as shown in Table~\ref{tab:osworld40_ablation}, suggesting that access to established progress helps avoid redundant recovery work.

\begin{table}[t]
\centering
\setlength{\abovecaptionskip}{10pt}
\setlength{\belowcaptionskip}{8pt}
\caption{\textbf{Context-view ablation on the shared 40-task subset of OSWorld 2.0.}
Each ablation hides one component from the host agent while preserving CarryOn's internal runtime state.
Removing either component substantially lowers partial score, while removing continuation carriers also increases agent calls and cost.}
\label{tab:osworld40_ablation}

\small
\renewcommand{\arraystretch}{1.12}
\setlength{\tabcolsep}{11pt}

\begin{tabularx}{1\linewidth}{>{\raggedright\arraybackslash}Xcccc}
\toprule
Method & Agent Calls & Cost (\$) $\downarrow$ & Binary (\%) $\uparrow$ & Partial (\%) $\uparrow$ \\
\midrule
\addlinespace[3pt]
w/o Execution Context &
274.33 & 18.33 & 17.50 & 44.00 \\
w/o Continuation Carriers &
349.45 & 25.14 & \textbf{25.00} & 44.62 \\
\cmidrule{1-5}
\textbf{CarryOn (ours)} &
282.25 & 20.55 & 22.50 & \textbf{55.93} \\
\bottomrule
\end{tabularx}
\end{table}

\label{sec:exp}

%% file: sec/5_conclusion.tex
\section{Conclusion}
We introduce CarryOn, a context runtime that organizes long-horizon agent context around reusable task objects. Our method combines continuation carriers that expose reusable progress, execution context that preserves complementary goals, constraints, and unresolved issues, and a live tail of recent interactions. As the task progresses, CarryOn jointly updates continuation carriers and execution context to reflect evolving task state and remaining requirements, progressively replacing covered interaction history with an updated context view. On OSWorld 2.0, CarryOn raises partial score from 43.9\% to 49.7\% while reducing total token usage by 51.9\% and nominal model cost by 35.9\%, including runtime overhead. On BrowseComp-Plus, it increases accuracy from 49\% to 56\% with lower token usage and cost. Together, these results support object-oriented context management for long-horizon agents, where task objects carry established progress and explicit context evolves with what is still needed to move forward.
\label{sec:con}